\documentclass[conference]{IEEEtran}
\IEEEoverridecommandlockouts
\usepackage{cite}
\usepackage{amsmath,amssymb,amsfonts}
\usepackage{algorithmic}
\usepackage{graphicx}
\usepackage{textcomp}
\usepackage{xcolor}
\usepackage{mathtools}

\def\BibTeX{{\rm B\kern-.05em{\sc i\kern-.025em b}\kern-.08em
    T\kern-.1667em\lower.7ex\hbox{E}\kern-.125emX}}
\begin{document}

\makeatletter
\newcommand{\linebreakand}{%
  \end{@IEEEauthorhalign}
  \hfill\mbox{}\par
  \mbox{}\hfill\begin{@IEEEauthorhalign}
}
\makeatother

\graphicspath{ {figures/} }

\IEEEoverridecommandlockouts
\IEEEpubid{\makebox[\columnwidth]{978-1-6654-0126-5/21/\$31.00~\copyright2021 IEEE \hfill} \hspace{\columnsep}\makebox[\columnwidth]{ }}

\title{Understanding the factors related to the opioid epidemic using machine learning}

\author{\IEEEauthorblockN{Sachin Gavali}
\IEEEauthorblockA{
\textit{University of Delaware}\\
Newark, DE, USA \\
saching@udel.edu}
\and
\IEEEauthorblockN{Chuming Chen}
\IEEEauthorblockA{
\textit{University of Delaware}\\
Newark, DE, USA \\
chenc@udel.edu}
\and
\IEEEauthorblockN{Julie Cowart}
\IEEEauthorblockA{
\textit{University of Delaware}\\
Newark, DE, USA \\
jcowart@udel.edu}
\linebreakand
\IEEEauthorblockN{Xi Peng}
\IEEEauthorblockA{
\textit{University of Delaware}\\
Newark, DE, USA \\
xipeng@udel.edu}
\and
\IEEEauthorblockN{Shanshan Ding}
\IEEEauthorblockA{
\textit{University of Delaware}\\
Newark, DE, USA \\
sding@udel.edu}
\and
\IEEEauthorblockN{Cathy Wu}
\IEEEauthorblockA{
\textit{University of Delaware}\\
Newark, DE, USA \\
wuc@udel.edu}
\linebreakand
\IEEEauthorblockN{Tammy Anderson}
\IEEEauthorblockA{
\textit{University of Delaware}\\
Newark, DE, USA \\
tammya@udel.edu}
}

\maketitle

\IEEEpubidadjcol

\begin{abstract}
    In recent years, the US has experienced an opioid epidemic with an unprecedented number of drugs overdose deaths.
    Research finds such overdose deaths are linked to neighborhood-level traits, thus providing opportunity to identify
    effective interventions.  Typically, techniques such as Ordinary Least Squares (OLS) or Maximum Likelihood
    Estimation (MLE) are used to document neighborhood-level factors significant in explaining such adverse outcomes.
    These techniques are, however, less equipped to ascertain non-linear relationships between confounding factors.
    Hence, in this study we apply machine learning based techniques to identify opioid risks of neighborhoods in
    Delaware and explore the correlation of these factors using Shapley Additive explanations (SHAP). We discovered
    that the factors related to neighborhoods’ environment, followed by education and then crime, were highly correlated
    with higher opioid risk. We also explored the change in these correlations over the years to understand the changing
    dynamics of the epidemic. Furthermore, we discovered that, as the epidemic has shifted from legal (i.e.,
    prescription opioids) to illegal (e.g., heroin and fentanyl) drugs  in recent years, the correlation of environment,
    crime and health related variables with the opioid risk has increased significantly while the correlation of
    economic and socio-demographic variables has decreased. The correlation of education related factors has been higher
    from the start and has increased slightly in recent years suggesting a need for increased awareness about the opioid
    epidemic.  
\end{abstract}


\section{Introduction}
For about 30 years, the United States (U.S.) has suffered a widespread crisis of opioid addiction and overdose deaths.
In 2019 alone, more than 50,000 Americans lost their lives to opioid overdose \cite{Abuse_2021}. This is an almost 200\%
increase in deaths compared to just a decade ago (16,651 people in 2010) \cite{Rudd_Seth_David_Scholl_2016}. 
Subsequently, there has been a significant effort to elucidate the factors driving this epidemic and enact strategies to
combat it. Led by the White House’s Office of National Drug Control Policy \cite{ndcb}, federal agencies such as the
Drug Enforcement Agency (DEA), the Centers for Disease Control and Prevention (CDC), the Department of Health and Social Services
(DHSS), and the National Institutes of Health (NIH) have taken several initiatives to address these problems
\cite{Soelberg_Brown_2017}. Despite these initiatives, there has been a consistent increase in deaths from both legal
and illegal opioids \cite{Abuse_2020, anderson_wagner}.
Traditionally techniques such as Ordinary Least Squares (OLS) and Maximum Likelihood Estimation (MLE) are not well
suited to exploit the increasingly non-linear relationships between various neighborhood-level indicators of opioid overdose
\cite{rf_vs_lr}. Machine learning techniques coupled with widely available neighborhood level
surveillance data can be a valuable tool to model these non-linear relations.  Hence, in this study we utilize the
Random Forest \cite{rf} to build a model to identify the opioid risk of neighborhoods across Delaware. We use
socio-demographic data from the US Census, crime data from the Delaware Criminal Justice Information System (DELJIS),
and opioid overdose data from Delaware Division of Forensic Science (DFS) \cite{domip}.
Understanding the factors driving the epidemic is very important in formulating effective intervention strategies.
Therefore, we use recent advancements in model interpretability to analyze our model \cite{SHAP} and understand the factors it
considers important in identifying the opioid risk. In addition to understanding the globally important
factors, we also compare the change in these factors over the years to understand the changing dynamics of the opioid
epidemic.

\section{Methods}

\subsection{Outcome variable — Opioid risk}

In this study our goal was to use machine learning methods to identify neighborhoods (i.e. census tracts) in Delaware
at an increased risk of overdose deaths and understand which neighborhood-level factors were correlated with them.
Delaware, a small state located in north-eastern part of the country, has been among the worst affected states by the
epidemic. It has consistently ranked in the top ten  states for highest overdose death rate, and in 2018 reported the
second-highest overdose death rate in the nation \cite{Abuse_2020}. With its mix of urban regions to the north and rural
regions to the south, Delaware is  a microcosm to discover insights that might be applicable to the larger U.S.
\cite{drug_situation}. 

For this study we built a longitudinal dataset of [census-tract + year] combinations containing 1444 observations for
overdose deaths  at census tract level from 2013 to 2019. We use the overdose death data from the Delaware Opioid Metric
Intelligence Project (DOMIP) \cite{domip}.

While substance abuse in the U.S. has increased in recent years, research shows it is still patterned by region
\cite{Cerda_2013}. This is true in Delaware as well, where some neighborhoods experience a higher level of
opioid-related overdose deaths compared to others [Figure \ref{fig:opioid_od_map}]. This results in data having  a
semicontinuous distribution with a large proportion of neighborhoods (\~50\%) having no deaths [Figure
\ref{fig:opioid_od_histogram}]. When studying data with a large proportion of zeros, one cannot use commonly used
methods such as linear regression as the resulting model will be biased towards zeros. Ideally, it is advised to adopt a
two-part model where we fit two models — A) Classification model to differentiate between zero and non-zero values, and
B) Regression model to model the death rate \cite{Boulton_Williford_2018}. Initially we adopted this strategy, and
though the model performance for the classification model was good (AU-PR = 0.65), the performance for the regression
model as measured using R2 score was not optimum (24\%).

\begin{figure}[htbp]
    \centerline{
        \scalebox{0.55}{
            \includegraphics[width=0.50\textwidth]{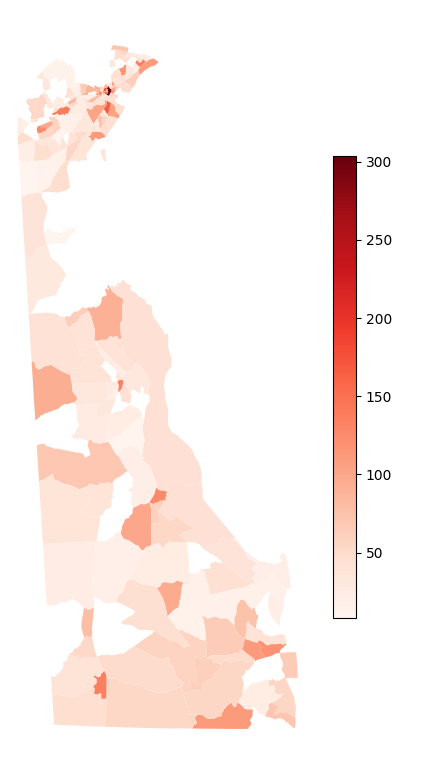} 
        }
    }
    \caption{Opioid related overdose deaths in 2019.}
    \label{fig:opioid_od_map}
\end{figure}

Since our goal in this study was to identify neighborhoods with an increased opioid risk, and not predict the death
rate, we decided to discretize the continuous outcome variable i.e., opioid overdose deaths to a categorical variable
named opioid risk having two categories — High and Low. In consultation with subject-matter experts, we decided to
specify any neighborhoods having opioid overdose death rate equal to or above 75th percentile to be in the High-risk
category and any neighborhoods having opioid overdose death rate less than 75th percentile to be in the Low-risk
category.

\begin{figure}[htbp]
    \centerline{
        \scalebox{0.75}{
            \includegraphics[width=0.50\textwidth]{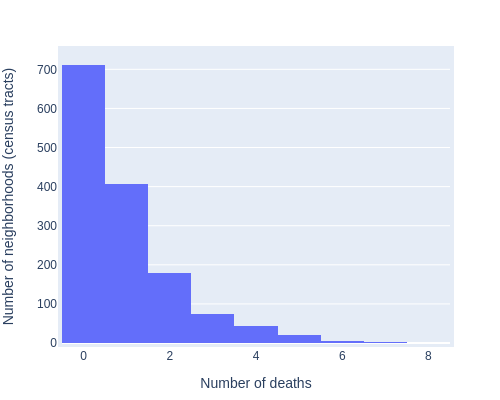} 
        }
    }
    \caption{Distribution of opioid related overdose deaths from 2013 to 2019.}
    \label{fig:opioid_od_histogram}
\end{figure}

\subsection{Determinants of opioid risk}
In recent years, research has found  social and environmental factors play an important role in determining the
health of individuals in a neighborhood \cite{Braveman_Gottlieb_2014}, including opioid consequences
\cite{revisiting_nbr,Anderson_Donnelly_Delcher_Wang_2021,gender_gap,pdh}. Such influential neighborhood-level
characteristics can include population density, racial and gender diversity, and measures of concentrated disadvantages
such as poverty or education \cite{revisiting_nbr, opioids_race,understanding_geo_nbr}.

Thus, informed by prior research \cite{Cerda_2013,Hall_2008,Modarai_2013,Li_Xu_Shah_Mackey_2019} we choose fifty-five
variables across six different categories — 1) Socio-demographic, 2) Environment, 3) Education, 4) Crime, 5) Economic,
and 6) Health to serve as determinants of the opioid risk of a neighborhood [Table \ref{table:predictors}].

We retrieved the data for all categories except the Crime category from the American Community Survey (ACS) conducted by
U.S. Census Bureau \cite{acs}. We retrieved the data at the census tract level for years 2013 to 2019. For the Crime
category, we use the arrests' data from Delaware Criminal Justice Information System (DELJIS) at the census tract level
from years 2013 to 2019.

\begin{table*}[htbp]
    \caption{Predictor variables}
    \begin{center}
    \scalebox{0.78}{
        \begin{tabular}{|p{0.2\linewidth} | p{0.18\linewidth} | p{0.18\linewidth}| p{0.18\linewidth}| p{0.18\linewidth}| p{0.18\linewidth}|}
        \hline
        \textbf{Sociodemographic } & \textbf{Environment} & \textbf{Education} & \textbf{Crime} & \textbf{Economic} & \textbf{Health}  \\
        \hline

        Population Density \newline 
        Male and Female population [under 18 years, 18 to 44 years, 45 to 65 years, 65 years and above] \newline
        Race [Black, White, Other] \newline
        Home Language [English, Non-English]
        & 
        Total Number of housing units \newline
        Number of occupied housing units \newline
        Number of vacant housing units \newline
        Number of households with no vehicle \newline
        Number of households with coal as primary source of heating \newline
        Number of households with no fuel for heating \newline
        Number of mobile homes \newline
        Number of female headed households \newline
        No vehicle for transport to work \newline
        No public transport vehicle for transport to work
        &
        Population with education below college \newline
        Population with college education, but no degree 
        &
        Population arrested for violent crimes \newline
        Population arrested for property crimes \newline
        Population arrested for socially important crimes \newline 
        Number of juveniles arrested for any crime \newline
        Number of youths arrested for any crime
        &
        Median income - [All, black, white, other] \newline
        Median rent \newline
        Total employed population [All sectors, agriculture and mining, construction, manufacturing, wholesale,
        transportation and warehousing, arts and recreation] \newline
        Population below poverty [All, white, black, other]
        &
        Disabled population [All, males, females] \newline
        Without health insurance [All, males, females, all below 18 years, all from 18 to 65 years, all above 65 years] \newline 
        \\
        \hline

        \end{tabular}
    }
    
    \label{table:predictors}
    \end{center}
\end{table*}

\subsection{Machine learning}

\subsubsection{Data preparation}
Since the ACS data is collected from surveys, it often contains missing values. A majority of the variables in our model
had less than 7.0\% missing values except for "Number of households with coal as the primary source of heating" and
"Number of Mobile Homes" which had 26.69\% and 36.92\% missing values respectively. Thus, we preprocessed the predictor
variables to replace missing values with their averages across the years. Then, we randomly split the data into training
(80\%) and testing (20\%) sets. The training set contained 1155 samples of which 866 samples belonged to low risk
category and 289 samples belonged to high risk category and testing set contained 289 samples of which 221 samples
belonged to low risk category and 68 samples belonged to high risk category. We used the training set for
hyper-parameter optimization and model training, and the testing set for evaluating the final model performance.

\subsubsection{Model training and evaluation}
Since we had limited samples for training the model we used K-Fold cross validation (k=10) to train and evaluate the
model on all possible splits of training data while performing the hyper-parameter search. Once we had the best
performing hyper-parameters, we trained the final model using these hyper-parameters and then evaluated its generalized
performance on the testing data.

Since our target variable was highly imbalanced, traditional metrics such as accuracy and area under receiver operating
characteristics (AU-ROC) would not provide an appropriate overview of the model performance. Therefore, we evaluated our
model using area under precision and recall curve (AU-PR). 

\subsubsection{Feature importance}
For the end users to have trust in the predictions, it is essential  to know the factors the model considers important
in making the predictions. Interpretability of machine learning models can be broadly achieved either by building
simpler models that are intrinsically explainable or by post hoc analysis after training the models. The first approach
sacrifices predictive performance in favor of interpretability but is not generalizable to any other types of models.
The second approach preserves the predictive performance but requires additional effort to interpret the model but is
more generalizable \cite{molnar}. In this study, we adopt the second approach and make use of the Shapley Additive
Explanations (SHAP) \cite{SHAP} to understand our model.

\subsection{Results and Discussion}

\subsubsection{Model performance}
We executed the hyper-parameter tuning step in the above pipeline 100 times, each time randomly changing the
hyper-parameters. After 100 iterations, we sorted the results by AU-PR to obtain the best performing hyper-parameters
and train the final model. Figure \ref{fig:au-pr} provides an overview of the performance of the best model [AU-PR
= 0.65].

\begin{figure}[htbp]
    
    \centerline{
        \scalebox{0.75}{
            \includegraphics[width=0.50\textwidth]{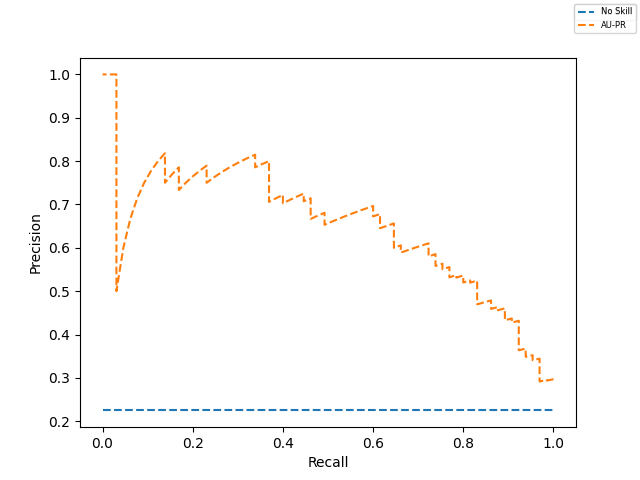}
        }
    }
    \caption{The area under PR-Curve (AU-PR) is an indicator of the model performance with
            higher AU-PR indicating better model performance. The no-skill line indicates the performance of a model 
            that predicts random classes regardless of the input and serves as a 
            baseline to validate the model.}
    \label{fig:au-pr}
\end{figure}

\subsubsection{Globally important community level factors}
In our pursuit to understand the factors influencing the overall opioid risk of neighborhoods, we decided to investigate
the factors influencing the predictions irrespective of the geographical space or the year involved. To do so, we
calculated the feature importance for every sample in the testing set. Shapley values can be either positive or negative
depending on the direction in which they help the model prediction. In our case, since our model is a binary
classification model, the direction of the Shapley value does not provide any meaningful information, but the magnitude
of the Shapley value provides an indication of the feature’s importance for the model. Hence, we convert all the feature
importance to their absolute values, and then we calculated an average of every feature across the 289 samples to get an
overview of the global feature importance. Table \ref{table:global-feature-importance} shows the top 10 features that
are globally correlated with an increased opioid risk of neighborhoods in Delaware from year 2013 to 2019.

\begin{table}[htbp]
    \caption{Globally important community level factors}
    \begin{center}
    \scalebox{0.8}{
        \begin{tabular}{|p{0.16\linewidth} | p{0.5\linewidth} | p{0.16\linewidth}|}
            \hline
            \textbf{Category} & \textbf{Feature} & \textbf{Importance} \\
            \hline
            Environment & Number of households with no vehicle & 0.045 \\
            \hline
            Environment & Number of households with coal as the primary source of heating & 0.029 \\
            \hline
            Education & Population with no college education & 0.022 \\
            \hline
            Education & Population with college education but no degree & 0.017 \\
            \hline
            Crime & Number of people arrested for property crimes & 0.014 \\
            \hline
            Health & Number of males with disability & 0.013 \\
            \hline
            Crime & Number of people arrested for violent crimes & 0.011 \\
            \hline
            Economy & Median Income & 0.008 \\
            \hline
            Health & Number of females with disability & 0.007 \\
            \hline
            Socio-demogrophic & Number of households with English as the only language & 0.007 \\
            \hline
            \end{tabular}
    }
    \label{table:global-feature-importance}
    \end{center}
\end{table}

The “number of households with no vehicles” and “number of households with coal as a primary source of heating” are
primarily the indicators of urbanity or rurality of a neighborhood. Households in urban regions usually have better
access to public transportation require less access to cars for daily activities \cite{public_transport}. On the other
hand, use of coal as a source of heating is more common in rural regions \cite{coal}. These correlations are in line
with existing research that has suggested that in recent years opioid epidemic has been expanding from rural regions to
urban regions \cite{Cerda_2017}. Higher levels of education have been associated with healthier populations
\cite{Zajacova_Lawrence_2018}. Hence, “Population with no college education” being one of the top factors is consistent
with past research. Prior studies have also shown opioid-related overdose mortality is significantly higher in adults
with a disability attributed to the misuse of opioids prescribed for pain relief \cite{opioid_pre_patterns}. This is
evident in our model were health related variables —“Number of males with disability” and “Number of females with
disability” are among the important features.  For socio-demographic variables, “Number of households with English as
the only language” and“White population” are congruent with the existing research that has suggested  the opioid
epidemic has primarily affected non-Hispanic white middle-aged male population \cite{King_2014}.

Our most interesting observation, though, is that crime- related variables have a significantly higher correlation with
the opioid risk than socio-demographic, health or economic variables. This could be an indication of the ongoing shift
in the nature of the epidemic from legal prescription opioids to illegal synthetic opioids, at least with respect to the
cause of overdose deaths.

Individual correlations of neighborhood factors with opioid risk, though very important, do not provide a complete
picture. Often, these individual neighborhood factors can be grouped into concepts/categories that embody a particular
facet of our communities. Hence, we also calculate aggregated feature importance according to these concepts or
categories. Figure \ref{fig:category-opioid-od} shows the relative importance of six categories in identifying the
opioid risks of neighborhoods in Delaware from year 2013 to 2019.

The environment is the most important factor in determining the overall opioid risk. It is followed by education and
crime related factors suggesting an ongoing change in the nature of the epidemic from legal to illegal drugs.
Socio-demographic and health related factors are equally important which might indicate that this epidemic has been
primarily affecting a particular subset of the population. 

\begin{figure}[htbp]
    \scalebox{0.85}{
        \centerline{\includegraphics[width=0.50\textwidth]{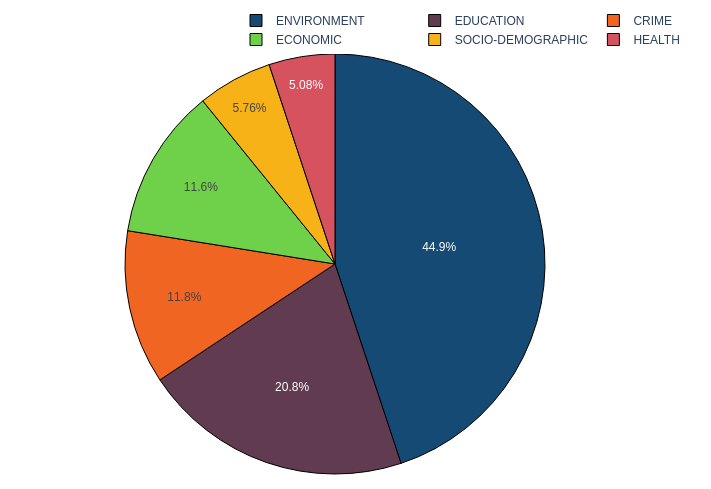}} 
    }
    \caption{Relative importance of different factors in determining the opioid risk.}
    \label{fig:category-opioid-od}
\end{figure}

\subsubsection{Change in the dynamics of the epidemic}
The Centers for Disease Control and Prevention (CDC) has characterized the U.S. opioid epidemic as having three distinct
waves, with the first wave starting in 1991 \cite{understanding_epidemic}. This first wave was primarily attributed to
the widespread use of prescription opioids, the second wave starting in 2010 was characterized by an increase in deaths
due to illicit opioids such as heroin and the third wave starting in 2013 can be attributed to Illegally manufactured
fentanyl — a highly potent synthetic opioid. Since then, fentanyl in combination with other drugs such as heroin and
cocaine have dominated opioid related overdose deaths [Figure \ref{fig:all-od}] \cite{understanding_epidemic}.

\begin{figure}[htbp]
    \centerline{
        \scalebox{0.85}{
            \includegraphics[width=0.50\textwidth]{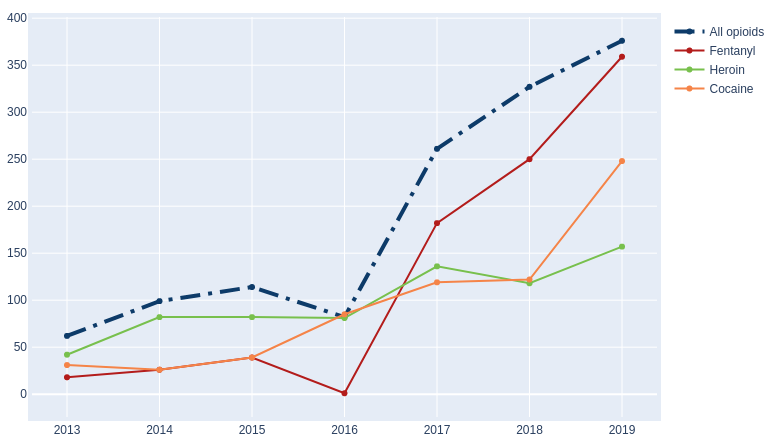}
        }
    }
    \caption{Overdose deaths in Delaware from 2013 to
    2019. Total number of overdose deaths have increased tremendously since 2016 and can be attributed primarily to
    fentanyl. Deaths due to cocaine have also been increasing steadily and have reached their highest levels in 2019.
    Deaths due to heroin have seen increase but not as tremendously as fentanyl or cocaine.}
    \label{fig:all-od}
\end{figure}

Even though these waves were primarily attributed to different types of opioids, the underlying neighborhood factors
driving each has been understudied and hence we try to understand if there were any differences in neighborhood factors
driving these waves. As evident from [Figure \ref{fig:all-od}], overdose deaths in Delaware  have shifted after 2016. 
Hence, to tease out the factors that might be driving these changes, we developed two distinct models. The first model
was developed to identify only heroin risk and the second model was developed to identify [fentanyl+cocaine] risk. We
kept the determinant neighborhood factors constant for both the models and trained them using the same technique used to
train the opioid risk model. The [fentanyl+cocaine] model had an AU-PR score of 0.67 and the heroin risk model had an
AU-PR score of 0.32. The difference in AU-PR scores of both the models can be attributed to the fact that fentanyl and
cocaine related deaths have increased at a much faster pace compared to heroin deaths leading to a stronger signal in
the fentanyl+cocaine data. 

We extracted the feature importance from both models using the same method used to extract the global feature importance
in the previous section. Table \ref{table:heroin-feature-importance} and Table \ref{table:fentanyl-feature-importance}
show the top 10 factors correlated with heroin risk and the fentanyl and cocaine risk, respectively. [Figure
\ref{fig:heroin-fentanyl}] shows the difference between the factors considered important by both models.

The most striking difference between the two models is the inversion of importance of environment, socio-demographic and
crime factors. For the [fentanyl+cocaine] model, the importance of environment-related variables have doubled compared
to the heroin model. Taking a look at [Table \ref{table:heroin-feature-importance}] and [Table
\ref{table:fentanyl-feature-importance}], we see “Households with no vehicles” is an important factor for both the
models, but “Households with coal as a primary source of heating” and “Number of mobile homes” is among the top 10
important features for [fentanyl+cocaine] model suggesting a relatively rural predominance of the fentanyl and cocaine
risk.

Similar to the environment factors, the importance of crime related factors has also increased significantly.
Proliferation of illegal drug markets might  explain the importance of the number of arrests related to property crimes
and violent offenses. Researchers have shown that violence is frequently a defining feature of illicit markets
\cite{Andreas_Wallman_2009}. The violence could be directed against the law enforcement officers trying to curb the
illegal trade or  against  other market actors in an effort to gain market dominance \cite{Friman_2009}. In either case,
increased importance of crime related variables might be suggestive of an active and thriving drug market in Delaware.
Another contrasting feature of the [fentanyl+cocaine] model is the higher importance of health related variables such as
“Number males with disability” and “Number of females with disability” which mirrors prior knowledge of opioid overdose
risk being high in patients prescribed opioids for pain management
\cite{Dickinson_2000,Kuo_Raji_Goodwin_2019}.

The importance of  socio-demographic factors has decreased significantly in recent years. The heroin risk model suggests
a higher predominance of heroin risk in the white population due to higher importance of “Home language English Only”and
“White Population”. This is in line with the existing research that has suggested presence of significant racial
diversity in the use of heroin with white population being more likely to use heroin and hispanic and black population
more likely to use cocaine \cite{Bernstein_2005}. The [fentanyl+cocaine] model considers other racial
groups (Asian, Alaskan natives and Hawaiian natives) to be more important than either white or black population (see
supplementary file) suggesting a change in the population that is being affected by the epidemic in recent years.

Overall, it is evident that in the recent years, the opioid epidemic has transcended the boundaries of race and income
and is affecting almost all parts of the population in some manner. Primary drivers for the opioid epidemic in recent
years include variables related to the environment, crime and health as evident by their increased importance in
determining fentanyl and cocaine risk. Thus, interventions targeted at addressing these factors might inform best
practices in combatting the opioid epidemic.

\begin{figure}[htbp]
    \centerline{
        \scalebox{0.9}{
            \includegraphics[width=0.48\textwidth]{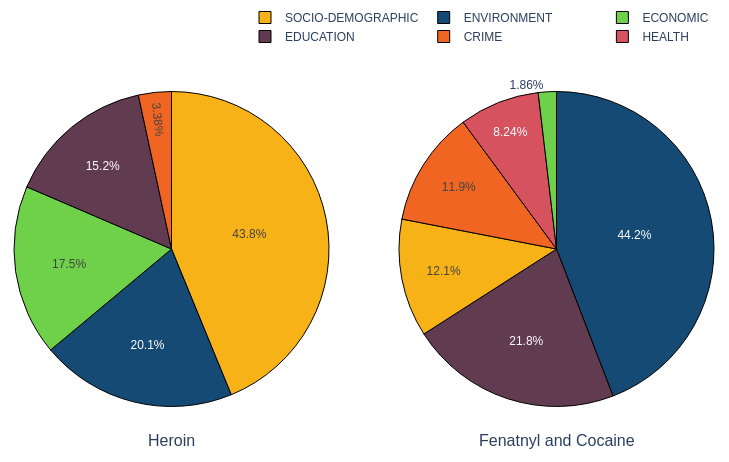}
        }
    } \caption{Relative
    importance of factors in identifying Heroin risk vs Fentanyl and Cocaine risk.}
    \label{fig:heroin-fentanyl}
\end{figure}

\begin{table}[htbp]
    \caption{Importance factors — Heroin risk}
    \begin{center}
    \scalebox{0.9} {
        \begin{tabular}{|p{0.16\linewidth} | p{0.5\linewidth} | p{0.16\linewidth} | }
        \hline
        \textbf{Category} & \textbf{Feature} & \textbf{Importance}  \\
        \hline
        Socio-demographic & Home language English only & 0.028  \\
        \hline
        Education & Population with no college education & 0.015  \\
        \hline
        Environment & Number of households with no vehicles & 0.014 \\
        \hline
        Socio-demographic & White population & 0.010 \\
        \hline
        Socio-demographic & Population density & 0.009 \\
        \hline
        Education & Number of people with college education but no degree & 0.009 \\
        \hline
        Environment & Total number of housing units & 0.06\\
        \hline
        Economy & Median income of white population & 0.008 \\
        \hline
        Environment & Number of occupied housing units & 0.007 \\
        \hline
        Economy & Number of people employed in construction & 0.007 \\
        \hline
        \end{tabular}
    }
    \label{table:heroin-feature-importance}
    \end{center}
\end{table}

\begin{table}[htbp]
    \caption{Importance factors — Fentanyl and Cocaine risk}
    \begin{center}
    \scalebox{0.9} {
        \begin{tabular}{|p{0.16\linewidth} | p{0.5\linewidth} | p{0.16\linewidth} |}
            \hline
            \textbf{Category} & \textbf{Feature} & \textbf{Importance}\\
            \hline
            Environment & Number of households with no vehicle & 0.059 \\
            \hline
            Education & Population with no college education & 0.033 \\
            \hline
            Environment & Number of households with coal as primary source of heating & 0.028 \\
            \hline
            Education & Number of people with college education but no degree & 0.021 \\
            \hline
            Crime & Number of people arrested for property crime & 0.016 \\
            \hline
            Crime & Number of people arrested for violent crime & 0.013 \\
            \hline
            Health & Male population with disability & 0.07 \\
            \hline
            Environment & Number of mobile homes & 0.011 \\
            \hline
            Health & Number of males with disability & 0.011 \\
            \hline
            Health & Number of females with disability & 0.009 \\
            \hline
            \end{tabular}
    }
    \label{table:fentanyl-feature-importance}
    \end{center}
\end{table}

\subsection{Conclusion}
In this work we have presented our approach of using machine learning to understand the factors driving the opioid
epidemic. We developed a model to identify the opioid risk of neighborhoods in Delaware using widely available
neighborhood level data. We used Shapely Additive Explanations to understand the correlation between neighborhood
factors and opioid related overdose deaths. Since the number of deaths increased dramatically after 2016, we developed
two additional models to understand the factors related to this dramatic increase in deaths.

We learned that in the recent years, environment related factors were the most significant drivers of the opioid
epidemic. This was followed by a significant increase in importance of crime related variables, suggesting an
accelerating shift in the opioid epidemic from the legal to illegal drugs. We also determined that, in accordance with
the existing knowledge, health related factors continue to be a significant determinant of opioid risk. Finally,
education related factors being equally important in determining heroin as well as fentanyl and cocaine risk suggests
that raising public awareness about the adverse effects of substance abuse might result in a decrease in mortality
due to substance abuse.

Machine learning models though tremendously useful in determining the correlation among the predictor and target
variables, are not appropriate for explaining the causal link between these variables. Therefore, the results from our
study should not be used as direct evidence of the causality of certain factors in determining opioid related overdose
deaths. Instead, we think our results can serve as a guide for future studies performing an in-depth investigation into
the casual relationships between these factors and the opioid epidemic.

\subsection{Acknowledgments}
We would like to thank Dr. Ellen Donnelly, Dr. Jascha Wagner, Andrew Gray, Logan Neitzke-Spruill and Cresean Hughes for
their helpful feedback on earlier drafts of this manuscript. Funding for this paper was provided by the National
Institute of Justice (NIJ), US Department of Justice, “Delaware Opioid Metric Intelligence Project,” (2017-IJ-CX-0016),
Tammy L. Anderson (Principal Investigator) and Daniel O’Connell (Co-Investigator). All findings and positions reported
here are those of the authors alone and do not reflect official positions of NIJ.

\bibliographystyle{IEEEtran}
\bibliography{domip_manuscript}

\end{document}